\documentclass[journal]{IEEEtran}
\usepackage{amsmath,amsfonts}
\usepackage{algorithmic}
\usepackage{algorithm}
\usepackage{array}
\usepackage{textcomp}
\usepackage{stfloats}
\usepackage{url}
\usepackage{verbatim}
\usepackage{graphicx}
\usepackage{cite}
\usepackage{color}
\usepackage{booktabs}
\usepackage{makecell}
\usepackage{multirow}
\begin{document}
\title{Federated Customization of Large Models: Approaches, Experiments, and Insights}
\author{{Yuchuan Ye, Ming Ding, Youjia Chen, Peng Cheng, and Dusit Niyato}
\thanks{\textit{Yuchuan Ye and Youjia~Chen (corresponding author) are with the College of Physics and Information Engineering, Fuzhou University, Fuzhou, China. Ming Ding is with Data61, CSIRO, NSW, Australia.
Peng Cheng is with the Department of Computer Science and Information Technology, La Trobe University, Melbourne, VIC, Australia. 
Dusit Niyato is with the College of Computing and Data Science, Nanyang Technological University, Singapore.
}}}

\maketitle

\begin{abstract}
In this article, we explore federated customization of large models and highlight the key challenges it poses within the federated learning framework.
We review several popular large model customization techniques, including full fine-tuning, efficient fine-tuning, prompt engineering, prefix-tuning, knowledge distillation, and retrieval-augmented generation. Then, we discuss how these techniques can be implemented within the federated learning framework. Moreover, we conduct experiments on federated prefix-tuning, which, to the best of our knowledge, is the first trial to apply prefix-tuning in the federated learning setting. The conducted experiments validate its feasibility with performance close to centralized approaches. Further comparison with three other federated customization methods demonstrated its competitive performance, satisfactory efficiency, and consistent robustness.
\end{abstract}


\section{Introduction}
\IEEEPARstart{I}{n} recent years, 
large models (LMs) have shown exceptional abilities in natural language processing and computer vision. With billions of parameters, they capture complex patterns and nuanced representations. Examples include OpenAI’s GPT-3 (175B) and Google’s PaLM (540B), both transformer-based models that set benchmarks in language understanding and reasoning. Vision transformers (ViTs) excel at image recognition, while multi-modal models like CLIP integrate text and images. Training such LMs from scratch is extremely resource-intensive and time-consuming and requires massive datasets: GPT-3 was trained on 570 GB of text (around 300 billion tokens) and PaLM on over 780 billion tokens. 

While general-purpose LMs demonstrate considerable capability, they frequently fall short when applied to specialized tasks. A more feasible and effective strategy is to leverage a pre-trained foundation LM as a starting point and subsequently adapt it through task- or domain-specific customization, that is, an efficient adaptation process in which a pre-trained model is refined for a specific downstream task without retraining the entire model from scratch.

Common customization methods generally include full fine-tuning of all parameters, efficient fine-tuning of a subset of parameters~\cite{adaptive_tuning,LoRA}, prompt engineering to optimize input prompts~\cite{in-context}, retrieval-augmented generation (RAG) that integrates external knowledge~\cite{RAG}, knowledge distillation transferring knowledge from an LM to smaller models~\cite{Task-Specific-KD}, and prefix tuning to prepend input prefixes~\cite{prefix-tuning}.

Effective LM customization often relies on task-specific sensitive data that cannot be centralized due to privacy and regulatory restrictions. For example, multiple hospitals may wish to collaboratively adapt a foundation LM with their local clinical records, but healthcare regulations prohibit data sharing. A similar challenge arises when enterprises seek to customize LM with private documents. Besides compliance concerns, aggregating such massive distributed datasets at a central server would also entail prohibitive communication overhead. In this context, federated customization of LM offers a practical and necessary solution.

Federated customization adapts the paradigm of federated learning (FL) to LMs, enabling collaborative training while keeping sensitive data decentralized. Instead of centralizing data, clients locally fine-tune task- or domain-specific components and share only updates with a central server, which aggregates them into a global model. This approach addresses privacy and regulatory constraints while lowering communication and computation by updating only essential components, offering a scalable, privacy-preserving, and efficient solution for deploying LMs across institutions and enterprises.

In this article, 
we first discuss the challenges and limitations of using FL for customizing foundation LMs. 
Then we review the popular customization techniques, 
including prompt engineering, full/efficient fine-tuning, prefix tuning, knowledge distillation, and RAG, and explore how they can be effectively adapted within the FL framework. 
Moreover, we conduct experiments on federated prefix-tuning, since, to the best of our knowledge, this is the first trial. 
Lastly, we outline potential future research directions for enabling and improving federated LM customization.

\section{Two Fundamental Aspects for Federated LMs}
In this section, we evaluate the feasibility of applying FL to LMs, outlining its challenges and identifying cases where it may be unsuitable.

\subsection{Federated Training of Foundation LMs}
Training a foundation LM in the FL framework 
allows access to the private data of multiple clients without data centralization, 
and enables distributed learning to alleviate the training burden of the central server. 
However, in practice, it faces enormous challenges and has the following drawbacks.

\subsubsection{Requirements on Training Data}

Foundational models such as GPT-3 require vast training data—often from hundreds of billions to trillions of tokens—to learn complex patterns and meanings behind them.
Currently, only large companies or institutions manage such massive and diverse datasets, making it challenging to collect this scale and variety of data through localized and individual clients.

Furthermore, 
foundational models need diverse datasets that can be applied to different situations, 
while FL usually uses specific, private data unique to each client. 
This lack of broad and varied data in FL limits the model's ability to learn widely useful patterns, making FL unsuitable for training foundational models.

\subsubsection{Higher Costs and Complexity}

The massive parameters in an LM require enormous computational capacity and memory size on the training device. 
Unlike centralized setups, 
where updates are efficiently managed within a single data center, FL requires each client to train a complete copy of the fundamental LM, 
imposing high memory demands on all clients. 
Moreover, 
FL requires frequent model aggregation and distribution between clients and the central server, 
which introduces extra communication costs. 
Hence, federated learning is actually more costly than centralized training.

\subsection{FL with LMs Is Unnecessary for Simple Tasks}
While FL is a compelling framework for privacy-preserving and decentralized training, applying it to LM is not always necessary, particularly for simple classification or regression tasks. LMs are designed for complex tasks such as language understanding and reasoning, making them computationally expensive. In contrast, simple tasks can often be solved effectively by lightweight models, rendering the use of FL with LMs excessive and inefficient in such scenarios.

For instance, 
in classification or regression tasks, 
the goal is to categorize input data or predict numerical values.
Smaller but task-specific models can handle these tasks effectively without the unnecessary complexity of an LM. In~\cite{vacareanu2024wordsnumberslargelanguage}, the authors found that while LMs such as GPT-4 can outperform traditional methods on challenging regression datasets, the performance gain is marginal. For instance, for the Liver Disorders dataset, GPT-4 achieved a mean absolute error of 2.55, slightly better than Gradient Boosting at 2.57. Considering the cost of LMs, traditional methods are a more practical choice for such tasks.

Furthermore, in~\cite{pawar2024generate}, the authors proposed a novel two-step technique, using a pre-trained LM followed by a lightweight classifier. Compared with fine-tuning LM, their approach achieved significant performance gains on text classification, with accuracy improvements of 0.273 on the SST-2 dataset and 0.07 on the TREC dataset. They concluded that the need for parameter updates in LM fine-tuning is eliminated, since a lightweight downstream model is sufficient enough for the classification tasks.

\section{LM Customization Techniques}
This section introduces key techniques for LM customization, grouped into {\emph{i)}} unfreezing methods (full and efficient fine-tuning) and {\emph{ii)}} frozen-model approaches (prompting, prefix-tuning, RAG, distillation).

\subsection{Full Fine-Tuning}
Full fine-tuning updates an entire LM model to improve its performance on a particular task. 
That is, 
all of the parameters of a pre-trained LM, 
such as GPT-3 or BERT, 
are retrained by the datasets related to the task of interest during the fine-tuning process. 
This approach undoubtedly yields strong performance due to its comprehensive adjustment of the model’s parameters, often outperforming the other LM customization methods \cite{10298587}.

However, 
full fine-tuning is the most resource-intensive. 
For instance, 
GPT-3 consists of 175 billion parameters, 
and new LMs tend to have even more parameters for better performance. 
Hence, fine-tuning demands significant computational resources,
including extensive graphics processing unit (GPU) memory, processing power, 
and large datasets. 
Specialized hardware, 
such as multi-GPU setups or TPUs, 
is generally required to handle the scale of these models efficiently.

\subsection{Efficient Fine-Tuning}
In contrast to full fine-tuning, 
which updates all the parameters of an LM, 
efficient fine-tuning keeps most of the LM frozen and only focuses on updating or adding a small part of the model, 
usually involving less than $5$ percent parameters. A straightforward approach to select tuning parameters is to use a sparse matrix, where only a small subset of the model's parameters are selected for fine-tuning. 

Another method is adapter-based tuning, 
in which lightweight neural network layers (termed adapters) are inserted between the layers of the transformer in the pre-trained LM~\cite{adaptive_tuning}. 
The experiment results show a comparable performance to full fine-tuning, 
particularly when training data is limited. 
 
Low-rank adaptation (LoRA) proposed in~\cite{LoRA} introduces small, task-specific matrices to a subset of the dense layers in the pre-trained LM. 
Specifically, it decomposes the weight update for certain layers into two lower-rank matrices, 
reducing the number of tunable parameters. 
These low-rank matrices are added to key weight matrices, 
such as those in the self-attention layers of transformer architectures, 
which optimizes memory usage and computational overhead. 

Low computational costs make it a more accessible solution for model customization. 
Even consumer-grade hardware, 
such as high-end laptops or standard cloud instances, 
is competent for efficient fine-tuning. 
However, 
for highly complex tasks, 
their performance may still fall short of full fine-tuning. 
Balancing efficiency and accuracy often requires carefully choosing detailed strategies, 
such as the proportion of fine-tuned parameters, ranks of decomposition matrices, and so on.

\subsection{Prompt Engineering}\label{Prompt Engineering}
Prompt engineering guides the pre-trained model's output through carefully designed input prompts. 
Generally, the input prompts provide contextual instructions or detailed examples to LMs, 
aiming to affect their generative capabilities.

One typical approach is in-context/few-shot learning \cite{in-context}, 
where a few task-related examples are included in the prompt to demonstrate the desired outcome, 
making the model efficiently adapt to new tasks. 
For instance, 
given a text translation task, 
a few sentences-pairs from a source language to a target language are provided, 
which allows the model to infer the translation pattern. 

Prompt engineering is highly flexible, 
allowing rapid experimentation across many applications, 
such as text generation/translation and question answering. 
Most importantly, 
it is free of model training, 
which implies no extra computing is required. 

However, prompt engineering is constrained by the LM’s token limit, 
i.e. the length of context in a single prompt. Also, it is limited by the capability of the model itself.
More importantly, 
crafting effective prompts within the token limit is not trivial——it requires a deep understanding of both the task and the LM's behavior. 
The quality of the prompt directly impacts the relevance and accuracy of the LM's outputs.

\subsection{Prefix-Tuning}\label{prefix-tuning}
Prefix-tuning \cite{prefix-tuning}, 
a parameter-efficient method, 
introduces an additional learnable vector, 
referred to as a "prefix", 
into the input sequence. 
These prefixes act as task-specific instructions that guide the LM’s behavior during inference. 

Generally,
these prefixes are optimized by a relatively small neural network trained on task-specific data. 
In \cite{prefix-tuning}, 
the prefix-optimizing network first embeds the input tokens, 
and then applies a simple network such as multi-layer perceptron (MLP) to produce vectors that match the dimensions of the transformer's key and value matrices. 
The prefix length serves as a tunable hyperparameter to control task-specific adaptation. 
Given an LM such as GPT, 
the optimized prefix vector is inserted directly into the transformer's key and value matrices, 
which adjusts the attention mechanism to the specific task.  

Prefix-tuning is highly efficient, 
but its performance is limited by the size of the prefix-optimizing network and the length of the optimized prefix. 
The design of prefix vectors is the key to task performance, 
which is a non-trivial work for complex tasks.

\subsection{Retrieval-Augmented Generation}
Retrieval-augmented generation (RAG) \cite{RAG} combines the strengths of retrieval-based methods and generative models to enhance the performance of LMs for specific tasks. 
It augments the LM's input with external, task-specific information retrieved from a knowledge base. 
By incorporating relevant context retrieved, RAG improves the quality, relevance, and accuracy of the content generated by LMs, especially for knowledge-intensive tasks.

RAG operates in two stages: retrieval and generation. 
In the retrieval stage, a query is sent to a retrieval tool/model, which searches for relevant information in an external database (e.g., a collection of task-specific documents or a knowledge graph). 
In the generation stage, the retrieved information is concatenated with the query and passed into a generative LM (such as GPT or BERT) to produce a task-oriented output. 

RAG allows the LM to access a broader knowledge base than what is stored in its parameters, 
effectively expanding the LM's ability to answer complex or domain-specific questions. 
On the other hand, the performance of a RAG system heavily depends on the quality and relevance of the retrieved documents. 
Hence, selecting an appropriate knowledge base and optimizing the retrieval model are crucial for obtaining high-quality outputs.

\subsection{Knowledge Distillation}\label{Knowledge Distillation}
Knowledge distillation aims to transfer knowledge from a large, pre-trained model (the teacher) to a smaller, more efficient model (the student). 
In this process, the teacher model generates soft labels (probabilistic outputs) or intermediate features (hidden layer activations), which guide the training of the student model by minimizing the loss function that combines the student’s own output and the teacher’s generated outputs.
 
For example, in~\cite{Task-Specific-KD}, for the sentence-pair task, the authors distill knowledge from BERT to generate a single-layer BiLSTM. The BiLSTM learns to minimize the mean-squared error between its logits and those of BERT, achieving competitive results with fewer parameters (100x fewer) and lower inference time (15x faster), matching the performance of ELMo (another widely used deep contextualized word representation model).

Knowledge distillation offers significant advantages in LM customization by reducing the resource cost, since it directly uses the learned knowledge from an LM. The main challenge for knowledge distillation is the detailed design of the student model, which can minimize the performance gap between the teacher and the student.

In summary, full fine-tuning offers strong performance but is resource-intensive. Efficient fine-tuning reduces costs but may underperform on complex tasks. Prompt engineering and prefix-tuning enable fast adaptation without retraining, though they are limited by token length or prefix design. RAG leverages external knowledge to improve output quality, depending on the relevance of retrieved content. Knowledge distillation yields smaller, efficient models with some performance trade-offs. The choice of approach depends on task complexity, resource constraints, and performance goals.

\section{Federated Customization of LMs}
In this section, we explore how the six LM customization techniques can be integrated into the FL framework, highlighting their potentials and limitations.

\begin{figure*}
    \centering
    \includegraphics[width=1\linewidth]{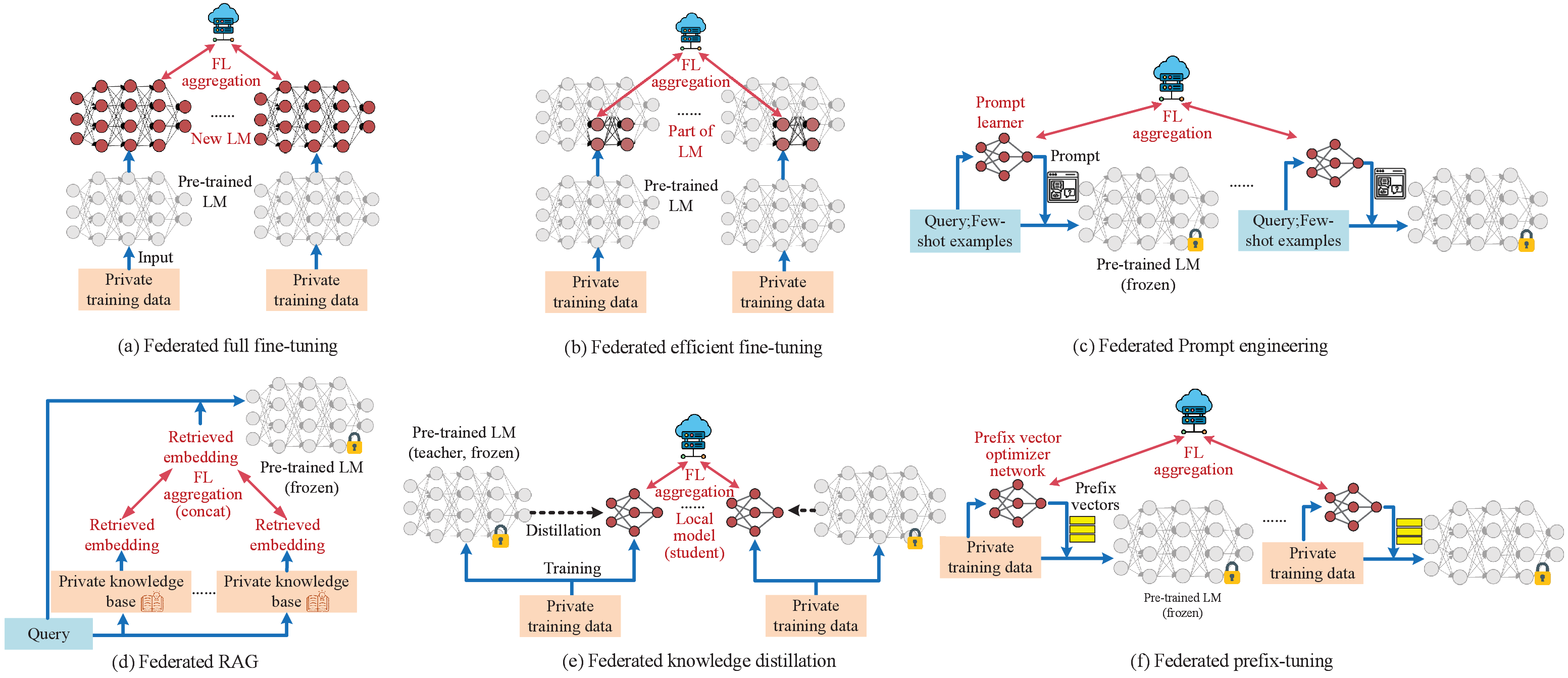}
    \caption{Frameworks of different federated LM customization methods.}
    \label{fig:fed_all_methods}
\end{figure*}

\subsection{Federated Full Fine-Tuning} 
As shown in Fig. \ref{fig:fed_all_methods}(a), federated full fine-tuning is a straightforward approach that applies standard FL techniques across multiple clients adopting full fine-tuning. 
In each FL iteration, 
each client fine-tunes the entire LM with its local data and periodically sends updated model parameters to a central server.
The server aggregates these parameters following a certain strategy and then distributes the aggregated model to all clients for further training. 

Federated full fine-tuning incurs high computational and communication costs, as each client must retrain the entire LM and exchange large parameter updates. This approach demands substantial hardware resources and imposes heavy communication overhead, making it impractical for most edge devices. Moreover, transmitting full model gradients poses privacy risks, as they can potentially reveal sensitive client data.

\subsection{Federated Efficient Fine-Tuning}

As shown in Fig.~\ref{fig:fed_all_methods}(b), compared with federated full fine-tuning, where the entire model needs to be aggregated, federated efficient fine-tuning methods only require the aggregation of the small subset of parameters tuned. Hence, the aggregation cost and the communication cost for FL are significantly reduced.

The performance of federated adapter tuning was tested in \cite{compare_fedefficien_full}. 
Using the large ViT-B model as an example, 
the communication cost for federated adapter tuning was reduced from 2.56 GBytes for full fine-tuning to 7.02 MBytes. 
On the CIFAR-100 dataset, 
federated full fine-tuning achieves an accuracy of $92.09$ percent, 
while federated adapter tuning can achieve $88.05$ percent. 
In~\cite{SLoRA}, 
the authors compared federated full fine-tuning with federated LoRA, 
showing that federated LoRA required only $5$ percent of the communication cost while achieving similar accuracy.

In summary, 
federated efficient fine-tuning methods offer a substantial reduction in communication and computation costs  
while achieving competitive performance with federated full fine-tuning. 
These advantages make federated efficient fine-tuning more feasible for LM customization in resource-sensitive scenarios.

\subsection{Federated Prompt Engineering}
As mentioned before, 
prompt engineering customizes models by constructing prompts customized to specific tasks or data contexts. 
However, directly aggregating these prompts to realize federated prompt engineering is inappropriate. 
As shown in Fig. \ref{fig:fed_all_methods}(c), federated prompt engineering generally deploys a prompt learner module on each client and then aggregates the prompt learners rather than the prompts themselves, 
such as PromptFL in \cite{promtfl}.

In PromptFL, 
each client employs a prompt learner to fine-tune continuous soft prompts, 
which are learnable embeddings represented by tunable vectors. 
In such method, the updates of the prompt learner model in each client are transmitted to the central server, where they are aggregated and redistributed for the next round of training until convergence. 
Performance evaluations show that PromptFL requires significantly fewer communication rounds, as little as 1.4 minutes to transfer 600 MB compared to FL’s nine hours for 40 GB.

\subsection{Federated RAG}
As shown in Fig. \ref{fig:fed_all_methods}(d), in a federated RAG system, 
data remains local to its source, 
and only necessary information, such as embedding vectors or model updates, is shared. 

A representative implementation is C-FedRAG~\cite{CFeDRAG}, which enables decentralized retrieval and embedding aggregation in federated RAG using confidential computing. Each data provider encodes its private corpus and returns relevant embeddings or text snippets in response to user queries. These are aggregated and re-ranked by a central orchestrator to construct an augmented query for LM inference. This design allows secure integration of distributed knowledge without requiring data centralization.

Another potential direction is training local retrievers at each client and aggregating their parameters in a federated manner. While this approach has been explored in the context of dense retrieval, integrating such mechanisms into federated RAG systems remains an open research area.

\begin{table*}[ht]
\centering
\caption{Comparison of federated LM customization methods.}
\resizebox{\textwidth}{!}{%
\begin{tabular}{|c|c|c|c|c|}
\hline
\textbf{Method} & \textbf{Feature}  & \textbf{Communication cost} & \textbf{Client-side computing cost} & \textbf{Server-side computing cost} \\ \hline
\makecell{\textbf{Federated} \\ \textbf{full fine-tuning}} & \makecell{Retrain \& aggregate \\ the entire LM} & High & High & Moderate \\ \hline
\makecell{\textbf{Federated} \\ \textbf{efficient fine-tuning}} & \makecell{Retrain \& aggregate \\ a portion of the LM} & \makecell{Low to moderate \\(depends on the portion)} & Moderate & Low \\ \hline
\makecell{\textbf{Federated} \\ \textbf{prompt engineering}} & \makecell{Train \& aggregate \\ prompt learners} & Low & Low & Low \\ \hline
\makecell{\textbf{Federated} \\ \textbf{prefix-tuning}} & \makecell{Train \& aggregate \\prefix optimizer models} & Low & Low & Low \\ \hline
\makecell{\textbf{Federated} \\ \textbf{RAG}} & \makecell{Aggregate embeddings; \\ train \& aggregate retrievers} & \makecell{Low to moderate\\(depends on embedding size)} & Low & High \\ \hline
\makecell{\textbf{Federated} \\ \textbf{knowledge distillation}} & \makecell{Train student models\\using the server-side LM} & \makecell{Low to moderate \\(depends on student model size)} & Low to moderate & Low \\ \hline
\end{tabular}%
}
\label{tab:fl_comparison}
\end{table*}

\subsection{Federated Knowledge Distillation}
As shown in Fig. \ref{fig:fed_all_methods}(e), in federated knowledge distillation, each client hosts a local teacher model (a pre-trained LM) and a student model (a smaller model). The teacher model distills its knowledge into the student model locally, and then the updates of the student model are uploaded to the server. The server aggregates these updates, creating a global student model.

In~\cite{wu2022communication}, each client distills knowledge from a local teacher to a student model using task loss (cross-entropy) and two adaptive distillation losses: mutual distillation (aligning soft labels) and hidden loss (aligning intermediate features and attention). After local training, student updates are aggregated into a global model and redistributed. On the MIND dataset, their method with a four-layer student achieved 71.0 percent AUC, outperforming FedAvg fine-tuning of compressed BERT at 69.7 percent.

Federated knowledge distillation reduces communication and computation costs by training and transmitting only the smaller student models. However, its performance may suffer under data heterogeneity, as diverse local datasets complicate knowledge integration. Additionally, variations in student model architectures can hinder direct aggregation, requiring output-level alignment.

\subsection{Federated Prefix-Tuning}
To the best of our knowledge, there is no existing literature specifically addressing federated prefix-tuning. Therefore, we are the first to explore this approach and design our own method for it.

As shown in Fig. \ref{fig:fed_all_methods}(f), to implement federated prefix-tuning, we avoid directly aggregating the learned prefix vectors from participating clients. Because these vectors are high-dimensional and unstructured latent parameters, simple averaging lacks theoretical justification. Instead, we aggregate the parameters of the prefix optimizer network, which defines a structured parameter space and thereby enables more effective and stable aggregation within the FL framework.
The proposed federated prefix-tuning consists of the following steps in each round:
\begin{itemize}
    \item Local Training of the Prefix Vector Optimizer: Keeping the LM frozen, each client uses private data to refine the prefix-related small neural network, such as a three-layer MLP. Then its parameters are sent to the central server.
    \item Model Aggregation and Distribution:
    The server aggregates received parameters by aggregation algorithms, such as FedAvg, creating a global model that captures the data features from all clients. Then, the aggregated global model is sent back to each client for the next round of local training. 
\end{itemize}

In Table \ref{tab:fl_comparison}, we summarize and compare the above federated LM customization methods, highlighting their costs on the computation and communication.

\section{Experiment Results}

\subsection{Experiment Setting}
We evaluate the table-to-text task on the E2E and DART datasets. The E2E dataset\footnote{https://github.com/tuetschek/e2e-dataset} contains about 42K examples, each consisting of structured meaning representations with eight fields (e.g., ‘name’, ‘food’, ‘price range’) paired with natural language descriptions. The DART dataset\footnote{https://github.com/Yale-LILY/dart} is larger, with roughly 82K examples derived from multiple sources, including WikiSQL, WikiTableQuestions, E2E, and DBpedia, and includes both manually and automatically generated text transformations. For evaluation, we adopt five official metrics—BLEU, NIST, METEOR, ROUGE-L, and CIDEr—where higher scores denote better performance. The prefix length is set to 10, following the configuration in~\cite{prefix-tuning}.

We consider 10 clients in FL, each holding one-tenth of the dataset and working on the same downstream task (table-to-text generation). Both IID and non-IID data partitions are evaluated.
Unless otherwise noted, GPT-2 Medium (GPT2-M) serves as the backbone LM, paired with a 25M-parameter MLP network for prefix optimization. Training employs early stopping if validation loss does not decrease for 3 consecutive epochs. In all experiments, each client completes one local epoch on its private data before every aggregation, which, based on prior theoretical studies \cite{li2019convergence} and our preliminary experiments, improves the global model’s performance by ensuring local updates contribute effectively to the aggregated model.

We compare four representative federated customization methods in our experiments: federated full fine-tuning (FFFT), federated adapter tuning (FAT), federated knowledge distillation (FKD), and federated prefix-tuning (FPT).
The classic FedAvg approach is employed to aggregate client updates in all methods. 

\subsection{Experiment Results}

\subsubsection{Feasibility Study of Federated Prefix-Tuning}

\begin{table*}
\caption{Performance of federated prefix-tuning across model scales and its comparison with centralized and single-client training.}
\centering
\label{result_table}
\setlength{\tabcolsep}{3pt}
\resizebox{\textwidth}{!}{
\begin{tabular}{llccccc|ccccc}
\toprule
& & \multicolumn{5}{c}{\textbf{E2E}} & \multicolumn{5}{c}{\textbf{DART}} \\
\cmidrule(lr){3-7} \cmidrule(lr){8-12}
& \textbf{Model} & \textbf{BLEU} & \textbf{NIST} & \textbf{METEOR} & \textbf{ROUGE-L} & \textbf{CIDEr}  
& \textbf{BLEU} & \textbf{NIST} & \textbf{METEOR} & \textbf{ROUGE-L} & \textbf{CIDEr} \\
\midrule
\multirow{3}{*}{FPT} & GPT-2 M (345M) & 68.91$\pm$0.12 & 8.80$\pm$0.02 & 46.25$\pm$0.21 & 71.71$\pm$0.08 & 2.48$\pm$0.01  
& 45.55$\pm$0.21 & 8.73$\pm$0.06 & 38.38$\pm$0.25 & 60.21$\pm$0.23 & 2.82$\pm$0.01 \\
& GPT-2 L (774M) & 69.98$\pm$0.15 & 8.82$\pm$0.13 & 46.43$\pm$0.14 & 71.45$\pm$0.19 & 2.51$\pm$0.11   
& 46.48$\pm$0.19 & 8.86$\pm$0.26 & 38.79$\pm$0.18 & 60.92$\pm$0.23 & 2.89$\pm$0.03  \\
& LLaMA-3.2 (1B) & 65.29$\pm$0.10 & 8.38$\pm$0.08 & 44.67$\pm$0.17 & 67.96$\pm$0.15 & 2.32$\pm$0.02  
& 40.24$\pm$0.21 & 8.18$\pm$0.09 & 37.37$\pm$0.18 & 55.82$\pm$0.22 & 2.48$\pm$0.02 \\
\midrule
\textbf{FPT}& GPT-2 M (345M) & \textbf{68.91$\pm$0.12} & \textbf{8.80$\pm$0.02} & \textbf{46.25$\pm$0.21} & \textbf{71.71$\pm$0.08} & \textbf{2.48$\pm$0.01}  
& \textbf{45.55$\pm$0.21} & \textbf{8.73$\pm$0.06} & \textbf{38.38$\pm$0.25} & \textbf{60.21$\pm$0.23} & \textbf{2.82$\pm$0.01}\\
\textbf{CPT} & GPT-2 M (345M) & \textbf{69.61$\pm$0.22} & \textbf{8.89$\pm$0.03} & \textbf{47.75$\pm$0.11} & \textbf{72.21$\pm$0.18} & \textbf{2.53$\pm$0.01}  
& \textbf{46.42$\pm$0.33} & \textbf{8.85$\pm$0.07} & \textbf{38.91$\pm$0.42} & \textbf{61.54$\pm$0.07} & \textbf{2.87$\pm$0.02}\\
\textbf{Client1-only} & GPT-2 M (345M) & 64.09$\pm$0.51 & 8.18$\pm$0.01 & 44.15$\pm$0.06 & 68.21$\pm$0.12 & 2.28$\pm$0.01  
& 40.78$\pm$0.36 & 8.06$\pm$0.12 & 36.45$\pm$0.50 & 56.82$\pm$0.18 & 2.47$\pm$0.02 \\
\textbf{Client2-only} & GPT-2 M (345M) & 66.24$\pm$0.24 & 8.44$\pm$0.09 & \textbf{45.27$\pm$0.27} & 68.80$\pm$0.56 & 2.35$\pm$0.01  
& 41.27$\pm$0.14 & \textbf{8.18$\pm$0.07} & 36.68$\pm$0.11 & 56.97$\pm$0.31 & 2.49$\pm$0.01 \\
\textbf{Client3-only} & GPT-2 M (345M) & 65.79$\pm$0.12 & 8.49$\pm$0.08 & 43.43$\pm$0.59 & 67.68$\pm$0.51 & 2.22$\pm$0.03  
& 41.28$\pm$0.26 & 8.11$\pm$0.05 & 36.51$\pm$0.28 & 56.76$\pm$0.03 & 2.47$\pm$0.01 \\
\textbf{Client4-only} & GPT-2 M (345M) & 63.93$\pm$0.11 & 8.29$\pm$0.07 & 43.63$\pm$0.36 & 67.68$\pm$0.54 & 2.18$\pm$0.04  
& 40.72$\pm$0.32 & 7.84$\pm$0.18 & 35.99$\pm$0.55 & \textbf{57.39$\pm$0.54} & 2.44$\pm$0.03 \\
\textbf{Client5-only} & GPT-2 M (345M) & 65.65$\pm$0.09 & 8.47$\pm$0.07 & 43.86$\pm$0.55 & 68.19$\pm$0.85 & 2.23$\pm$0.01  
& 41.17$\pm$0.09 & 8.02$\pm$0.07 & 36.19$\pm$0.45 & 56.94$\pm$0.08 & 2.47$\pm$0.01 \\
\textbf{Client6-only} & GPT-2 M (345M) & 63.61$\pm$0.32 & 8.12$\pm$0.06 & 42.58$\pm$0.19 & 68.01$\pm$0.16 & 2.10$\pm$0.02  
& 40.92$\pm$0.47 & 7.89$\pm$0.14 & 36.03$\pm$0.05 & 57.03$\pm$0.59 & 2.46$\pm$0.03 \\
\textbf{Client7-only} & GPT-2 M (345M) & 66.21$\pm$0.53 & 8.45$\pm$0.07 & 44.36$\pm$0.38 & 68.51$\pm$0.05 & 2.31$\pm$0.01  
& 40.39$\pm$0.39 & 7.75$\pm$0.06 & 35.91$\pm$0.36 & 56.47$\pm$0.54 & 2.41$\pm$0.01 \\
\textbf{Client8-only} & GPT-2 M (345M) & \textbf{67.08$\pm$0.47} & \textbf{8.61$\pm$0.04} & 44.32$\pm$0.45 & \textbf{69.05$\pm$0.02} & \textbf{2.36$\pm$0.02}  
& 39.67$\pm$0.62 & 7.85$\pm$0.13 & 35.87$\pm$0.11 & 56.11$\pm$0.05 & 2.44$\pm$0.01 \\
\textbf{Client9-only} & GPT-2 M (345M) & 64.76$\pm$0.45 & 8.32$\pm$0.13 & 43.41$\pm$0.11 & 68.11$\pm$0.08 & 2.25$\pm$0.01  
& 41.35$\pm$0.25 & 8.04$\pm$0.09 & \textbf{36.39$\pm$0.09} & 57.22$\pm$0.64 & \textbf{2.50$\pm$0.01} \\
\textbf{Client10-only} & GPT-2 M (345M) & 65.34$\pm$0.09 & 8.51$\pm$0.03 & 44.47$\pm$0.26 & 67.35$\pm$0.54 & 2.24$\pm$0.01  
& \textbf{41.46$\pm$0.84} & 8.15$\pm$0.05 & 36.25$\pm$0.66 & 57.13$\pm$0.74 & 2.50$\pm$0.01 \\
\bottomrule
\end{tabular}}
\end{table*}

\begin{table*}[ht]
\caption{Performance and resource comparison of different federated customization methods on E2E and DART datasets.}
\centering
\label{tab:compare_method_final}
\resizebox{\textwidth}{!}{
\begin{tabular}{l|l|ccccc|ccc}
\toprule
\textbf{Dataset} & \textbf{Method} 
& \textbf{BLEU} & \textbf{NIST} & \textbf{METEOR} & \textbf{ROUGE-L} & \textbf{CIDEr} 
& \makecell{\textbf{Trainable} \\ \textbf{Params (M)}} 
& \makecell{\textbf{Peak} \\ \textbf{Memory (GB)}} 
& \makecell{\textbf{Epochs} \\ \textbf{to Stop}} \\
\midrule
\multirow{4}{*}{E2E}
& FPT   & 68.91$\pm$0.12 & 8.80$\pm$0.02 & 46.25$\pm$0.21 & 71.71$\pm$0.08 & 2.48$\pm$0.01 & 25   & 4.8 & 17 \\
& FFFT  & 67.67$\pm$0.14 & 8.56$\pm$0.19 & 45.81$\pm$0.18 & 70.53$\pm$0.17 & 2.44$\pm$0.02 & 345  & 7.6 & 5  \\
& FAT  & 68.23$\pm$0.16 & 8.63$\pm$0.19 & 45.95$\pm$0.14 & 71.90$\pm$0.17 & 2.44$\pm$0.02 & 25  & 4.8 & 6  \\
& FKD   & 68.55$\pm$0.18 & 8.71$\pm$0.23 & 45.29$\pm$0.01 & 70.60$\pm$0.23 & 2.40$\pm$0.03 & 38.3 & 5.9 & 18 \\
\midrule
\multirow{4}{*}{DART}
& FPT   & 45.55$\pm$0.21 & 8.73$\pm$0.06 & 38.38$\pm$0.25 & 60.21$\pm$0.23 & 2.82$\pm$0.01 & 25   & 4.8 & 21 \\
& FFFT  & 34.19$\pm$0.28 & 6.50$\pm$0.25 & 38.70$\pm$0.19 & 55.00$\pm$0.17 & 1.95$\pm$0.01 & 345  & 7.6 & 6  \\
& FAT & 32.24$\pm$0.18 & 6.11$\pm$0.22 & 38.42$\pm$0.21 & 53.98$\pm$0.15 & 1.88$\pm$0.01 & 25 & 4.8 & 7  \\
& FKD   & 31.13$\pm$0.20 & 5.60$\pm$0.12 & 26.83$\pm$0.15 & 43.75$\pm$0.18 & 1.44$\pm$0.02 & 38.3 & 5.9 & 20 \\
\bottomrule
\end{tabular}
}
\end{table*}

To evaluate the effectiveness of FPT, we compare its performance with \emph{i)} different model scales under FPT, assessing scalability across GPT-2 M, GPT-2 Large (GPT-2 L), and LLaMA-3.2 1B, and \emph{ii)} centralized prefix-tuning (CPT) using the whole dataset, as well as independent client-only training, where each model is trained solely on the local training data of a single client and evaluated on the shared IID test set.

Table~\ref{result_table} presents the experimental results on the E2E and DART datasets. Firstly, the results show that FPT with GPT-2 Large achieves slightly higher scores than GPT-2 Medium on both datasets, demonstrating that the approach performs better as the model size increases within the GPT-2 family. In contrast, LLaMA-3.2 1B underperforms compared to both GPT-2 models, particularly on the DART dataset, likely because its pretraining data, which consists mainly of free-form natural text rather than structured table-to-text pairs, is less aligned with the benchmark tasks. 
Secondly, FPT consistently outperforms individual client training, demonstrating its effectiveness in improving overall performance. Moreover, the performance gap between FPT and CPT is marginal, further highlighting the effectiveness of FPT.

\subsubsection{Comparison of Different Federated Customization Methods}

We compare the performance and resource costs of four federated customization methods: FPT, FFFT, FAT, and FKD. As mentioned, FPT trains a lightweight prefix optimizer network with 25M parameters. FFFT updates all 345M parameters of GPT-2 Medium. FAT inserts adapters with 25M trainable parameters, while FKD trains a compact student model with 38.3M parameters guided by GPT-2 Medium.

As shown in Table \ref{tab:compare_method_final}, on the E2E dataset, all methods achieve comparable performance, with FPT slightly ahead. On the more challenging DART dataset, FPT clearly leads, while FFFT shows moderate performance, FAT experiences a decline, and FKD performs the worst among them.

Regarding efficiency, although FFFT converges the fastest, it requires substantial resources due to the large number of parameters. FAT incurs similar computation cost as FPT, while converging quickly. FKD has more memory usage due to the new student model and converges more slowly. FPT maintains a moderate computational and communication cost. Therefore, the choice of method should be based on the specific requirements of the task and the available device resources, balancing performance and efficiency.

\subsubsection{Impacts of Client Numbers}

We evaluated the performance with 20, 30, and 50 clients (see Table~\ref{client_scaling_table}), keeping the total dataset constant and equally divided among the clients. As the number of clients increases, the data per client decreases, leading to performance degradation across all methods.

Notably, FPT experiences the smallest drop on both datasets, FAT shows a moderate decline, while FKD degrades the most, especially on the DART dataset. This is due to their different underlying designs. FPT aggregates the parameters of the prefix optimizers while keeping the pretrained LM frozen, allowing the strong generalization ability of the pretrained LM to drive final performance. In contrast, FKD aggregates a completely new student model instead of the frozen backbone, making it more sensitive to the quality of local training.

Moreover, advanced aggregation algorithms may help reduce this performance degradation. Our experiments with FPT using the FedProx algorithm show performance gains of around 0.2–0.8 percent on E2E and 0.4–1.0 percent on DART compared to FedAvg, across different client numbers.

\begin{table*}[ht]
\caption{Performance of different federated customization methods with varying numbers of clients.}
\centering
\label{client_scaling_table}
\setlength{\tabcolsep}{3pt}
\resizebox{\textwidth}{!}{
\begin{tabular}{llccccc|ccccc}
\toprule
& & \multicolumn{5}{c}{\textbf{E2E}} & \multicolumn{5}{c}{\textbf{DART}} \\
\cmidrule(lr){3-7} \cmidrule(lr){8-12}
\multirow{-2}{*}{\makecell{\textbf{Client}\\\textbf{number}}} & \textbf{Method} & \textbf{BLEU} & \textbf{NIST} & \textbf{METEOR} & \textbf{ROUGE-L} & \textbf{CIDEr}  
& \textbf{BLEU} & \textbf{NIST} & \textbf{METEOR} & \textbf{ROUGE-L} & \textbf{CIDEr} \\
\midrule
\multirow{3}{*}{\textbf{20}} 
& FPT & 68.50$\pm$0.03 & 8.72$\pm$0.01 & 45.88$\pm$0.06 & 71.28$\pm$0.09 & 2.45$\pm$0.01  
& 44.64$\pm$0.35 & 8.56$\pm$0.06 & 38.07$\pm$0.18 & 59.74$\pm$0.27 & 2.76$\pm$0.03 \\
& FAT & 67.82$\pm$0.21 & 8.58$\pm$0.12 & 45.73$\pm$0.17 & 70.85$\pm$0.11 & 2.43$\pm$0.02  
& 31.78$\pm$0.21 & 6.11$\pm$0.22 & 38.39$\pm$0.18 & 53.63$\pm$0.18 & 1.72$\pm$0.02 \\
& FKD & 53.52$\pm$0.30 & 7.52$\pm$0.08 & 35.36$\pm$0.25 & 58.25$\pm$0.19 & 1.20$\pm$0.03  
& 22.69$\pm$0.41 & 4.51$\pm$0.18 & 21.97$\pm$0.26 & 35.09$\pm$0.23 & 0.79$\pm$0.02 \\
\midrule
\multirow{3}{*}{\textbf{30}} 
& FPT & 68.12$\pm$0.32 & 8.68$\pm$0.01 & 45.59$\pm$0.13 & 70.72$\pm$0.22 & 2.41$\pm$0.01  
& 44.41$\pm$0.05 & 8.50$\pm$0.01 & 37.86$\pm$0.06 & 59.59$\pm$0.11 & 2.74$\pm$0.01 \\
& FAT & 67.46$\pm$0.22 & 8.53$\pm$0.09 & 45.57$\pm$0.20 & 70.07$\pm$0.13 & 2.40$\pm$0.02  
& 32.14$\pm$0.20 & 6.24$\pm$0.16 & 38.31$\pm$0.22 & 53.85$\pm$0.21 & 1.72$\pm$0.02 \\
& FKD & 53.09$\pm$0.29 & 7.40$\pm$0.07 & 34.42$\pm$0.27 & 57.66$\pm$0.20 & 1.31$\pm$0.03  
& 19.05$\pm$0.42 & 4.17$\pm$0.17 & 19.21$\pm$0.25 & 33.64$\pm$0.22 & 0.61$\pm$0.02 \\
\midrule
\multirow{3}{*}{\textbf{50}} 
& FPT & 67.53$\pm$0.21 & 8.58$\pm$0.02 & 44.63$\pm$0.13 & 69.36$\pm$0.14 & 2.35$\pm$0.01  
& 42.73$\pm$0.03 & 8.13$\pm$0.64 & 36.76$\pm$0.03 & 58.31$\pm$0.10 & 2.58$\pm$0.01 \\
& FAT & 66.56$\pm$0.22 & 8.43$\pm$0.06 & 45.19$\pm$0.16 & 69.48$\pm$0.18 & 2.37$\pm$0.02  
& 29.87$\pm$0.26 & 5.89$\pm$0.21 & 37.56$\pm$0.22 & 48.93$\pm$0.16 & 1.62$\pm$0.02 \\
& FKD & 52.24$\pm$0.31 & 7.42$\pm$0.08 & 35.12$\pm$0.26 & 58.07$\pm$0.21 & 1.14$\pm$0.03  
& 17.46$\pm$0.44 & 3.79$\pm$0.19 & 17.74$\pm$0.24 & 31.95$\pm$0.21 & 0.50$\pm$0.02 \\
\bottomrule
\end{tabular}}
\end{table*}

\begin{table*}[ht]
\caption{Performance of different federated customization methods under non-IID settings.}
\centering
\label{noniid_result_table}
\setlength{\tabcolsep}{3pt} 
\resizebox{\textwidth}{!}{
\begin{tabular}{llccccc|ccccc}
\toprule
& & \multicolumn{5}{c}{\textbf{E2E}} & \multicolumn{5}{c}{\textbf{DART}} \\
\cmidrule(lr){3-7} \cmidrule(lr){8-12}
\textbf{Setting} & \textbf{Method} & \textbf{BLEU} & \textbf{NIST} & \textbf{METEOR} & \textbf{ROUGE-L} & \textbf{CIDEr}
& \textbf{BLEU} & \textbf{NIST} & \textbf{METEOR} & \textbf{ROUGE-L} & \textbf{CIDEr} \\
\midrule
\multirow{3}{*}{\makecell{\textbf{Non-IID}\\ \textbf{(80\%)}}} 
& FPT & 65.79$\pm$0.40 & 8.41$\pm$0.06 & 43.35$\pm$0.96 & 68.67$\pm$0.57 & 2.25$\pm$0.07
& 44.36$\pm$0.34 & 8.49$\pm$0.08 & 37.87$\pm$0.16 & 59.59$\pm$0.25 & 2.74$\pm$0.02 \\
& FAT & 64.15$\pm$0.25 & 8.21$\pm$0.05 & 44.20$\pm$0.30 & 67.85$\pm$0.40 & 2.23$\pm$0.03
& 23.10$\pm$0.40 & 4.00$\pm$0.10 & 36.00$\pm$0.35 & 44.20$\pm$0.30 & 0.60$\pm$0.04 \\
& FKD & 67.55$\pm$0.28 & 8.56$\pm$0.05 & 42.96$\pm$0.34 & 68.64$\pm$0.38 & 2.16$\pm$0.03
& 29.08$\pm$0.40 & 5.19$\pm$0.15 & 25.99$\pm$0.31 & 42.37$\pm$0.29 & 1.30$\pm$0.04 \\
\midrule
\multirow{3}{*}{\makecell{\textbf{Non-IID}\\ \textbf{(60\%)}}} 
& FPT & 66.06$\pm$0.09 & 8.44$\pm$0.02 & 43.70$\pm$0.46 & 68.75$\pm$0.29 & 2.28$\pm$0.01
& 44.52$\pm$0.04 & 8.55$\pm$0.01 & 37.95$\pm$0.05 & 59.64$\pm$0.07 & 2.76$\pm$0.01 \\
& FAT & 64.80$\pm$0.20 & 8.32$\pm$0.04 & 44.00$\pm$0.28 & 68.20$\pm$0.32 & 2.25$\pm$0.02
& 24.60$\pm$0.35 & 5.00$\pm$0.08 & 36.80$\pm$0.30 & 46.80$\pm$0.30 & 0.62$\pm$0.03 \\
& FKD & 67.76$\pm$0.24 & 8.54$\pm$0.04 & 45.81$\pm$0.22 & 70.25$\pm$0.29 & 2.43$\pm$0.02
& 30.02$\pm$0.32 & 5.49$\pm$0.13 & 26.44$\pm$0.29 & 43.17$\pm$0.27 & 1.37$\pm$0.04 \\
\midrule
\multirow{3}{*}{\makecell{\textbf{Non-IID}\\ \textbf{(40\%)}}} 
& FPT & 67.29$\pm$0.39 & 8.55$\pm$0.01 & 44.02$\pm$0.64 & 69.43$\pm$0.25 & 2.32$\pm$0.08
& 44.81$\pm$0.05 & 8.57$\pm$0.14 & 38.04$\pm$0.04 & 59.83$\pm$0.05 & 2.77$\pm$0.01 \\
& FAT & 66.20$\pm$0.25 & 8.40$\pm$0.05 & 44.50$\pm$0.30 & 69.00$\pm$0.35 & 2.28$\pm$0.03
& 25.50$\pm$0.38 & 5.10$\pm$0.10 & 37.50$\pm$0.32 & 47.50$\pm$0.35 & 0.70$\pm$0.04 \\
& FKD & 68.27$\pm$0.22 & 8.64$\pm$0.05 & 45.98$\pm$0.20 & 70.75$\pm$0.27 & 2.42$\pm$0.02
& 30.74$\pm$0.41 & 5.54$\pm$0.15 & 26.06$\pm$0.30 & 43.26$\pm$0.32 & 1.48$\pm$0.05 \\
\bottomrule
\end{tabular}}
\end{table*}

\subsubsection{Impacts of Non-IID Data}
In the E2E dataset, we model non-IID data by assigning different dietary preferences to clients, that is, x\% of Client1’s data is Italian cuisine, while x\% of Client2’s is French. For the DART dataset, we categorize data by source (WikiSQL, WikiTableQuestions, E2E, DBpedia), with x\% of each client’s data drawn from one source and the rest sampled from others. We set x = 80\%, 60\%, 40\%, where a larger x indicates a stronger non-IID. All evaluation results are reported on a shared IID test set, measuring the performance of the globally aggregated model.

From Table~\ref{noniid_result_table}, on the E2E dataset, the three methods under the non-IID setting achieve similar performance, with FAT performing a little worse. On the DART dataset, FPT clearly outperforms FAT and FKD under the non-IID setting. Comparing the performance on two datasets, we find that FPT is the most stable method, while FKD is sensitive to the task.

Also, as the degree of non-IID data increases, all methods experience performance degradation across both datasets. On E2E, performance drops are consistent at about 3 percent when the non-IID level rises from 40\% to 80\%. On DART, the impact is minimal for FPT (0.6 percent) but substantially higher for FAT (12 percent) and moderate for FKD (3 percent). These results highlight FPT’s relative robustness. 

For FPT, the performance drop primarily stems from the misalignment of prefix vectors across clients under heterogeneous data, as reflected by the cosine similarity between local and aggregated prefix vectors. On DART, the similarity decreases slightly (0.842 to 0.834) as the non-IID level rises from 40\% to 80\%, indicating that prefix vectors remain largely aligned and explaining FPT’s robustness. This stability can be attributed to the LM’s strong generalization ability and the prefix optimizer’s capacity to capture task-level patterns shared across clients. 

Furthermore, advanced aggregation algorithms also help. In our test, under the 80\% non-IID scenario, FedProx yields an average improvement of 2 percent on E2E and 1.5 percent on DART compared to FedAvg.

\section{Future Research Directions}

\subsection{Federated Customization In Multi-Task or Multi-Modal Scenarios}
LMs can support diverse tasks and modalities, which opens opportunities but also challenges for federated customization. In such scenarios, the model must adapt to each client’s specific tasks and modalities while the global model needs to maintain generalization across them. Potential strategies may include task-based transfer learning, selective parameter sharing, and the co-design of clients' feature adaptation modules and the server's alignment module.

\subsection{Privacy-Enhanced Federated Customization}
Ensuring privacy in federated customization of LM is essential, as sensitive data is used for client-specific tasks. Differential privacy can be integrated into methods like federated RAG, efficient fine-tuning, and prompt engineering by adding calibrated noise to local data or updates, limiting the influence of individual data points and reducing leakage risk. However, this requires balancing privacy and model accuracy, as noise may affect output quality.

\subsection{Lightweight Federated Customization of LMs}
Most existing approaches to federated customization require each client to store the full model, which is impractical for resource-constrained devices such as mobile phones. A promising direction is to integrate parameter-efficient customization with parallelism or offloading strategies—for example, using tensor and pipeline parallelism or split learning between edge devices and the cloud.

\section{Conclusions}
In conclusion, our study presented federated customization as a viable use case for deploying LMs in FL. By presenting the disadvantages of federated foundational LM training such as high data and resource demands, we highlighted \textit{the effectiveness of federated customization of LM} using techniques such as prompt engineering, full/efficient fine-tuning, 
prefix-tuning, knowledge distillation,
and retrieval-augmented generation. 
The experiments conducted on table-to-text generation tasks substantiated the feasibility of the proposed FPT framework. Further comparisons with three alternative federated customization methods demonstrated its competitive performance, satisfactory efficiency, and consistent robustness.
Future research could aim to make federated customization more scalable, practical, and adaptable to more diverse tasks and deployment scenarios.


\bibliographystyle{IEEEtran}
\bibliography{cite}

\end{document}